# DeepWriterID: An End-to-end Online Text-independent Writer Identification System

Weixin Yang, Lianwen Jin*, Manfei Liu
College of Electronic and Information Engineering, South China University of Technology, Guangzhou, China
wxy1290@163.com, *lianwen.jin@gmail.com

*Abstract*—Owing to the rapid growth of touchscreen mobile terminals and pen-based interfaces, handwriting-based writer identification systems are attracting increasing attention for personal authentication and digital forensics. However, most studies on writer identification have not been satisfying because of the insufficiency of data and the difficulty of designing good features for various conditions of handwriting samples. Hence, we introduce an end-to-end system called DeepWriterID that employs a deep convolutional neural network (CNN) to address these problems. A key feature of DeepWriterID is a new method we are proposing, called DropSegment. It is designed to achieve data augmentation and to improve the generalized applicability of CNN. For sufficient feature representation, we further introduce path-signature feature maps to improve performance. Experiments were conducted on the NLPR handwriting database. Even though we only use pen-position information in the pen-down state of the given handwriting samples, we achieved new state-of-the-art identification rates of 95.72% for Chinese text and 98.51% for English text.

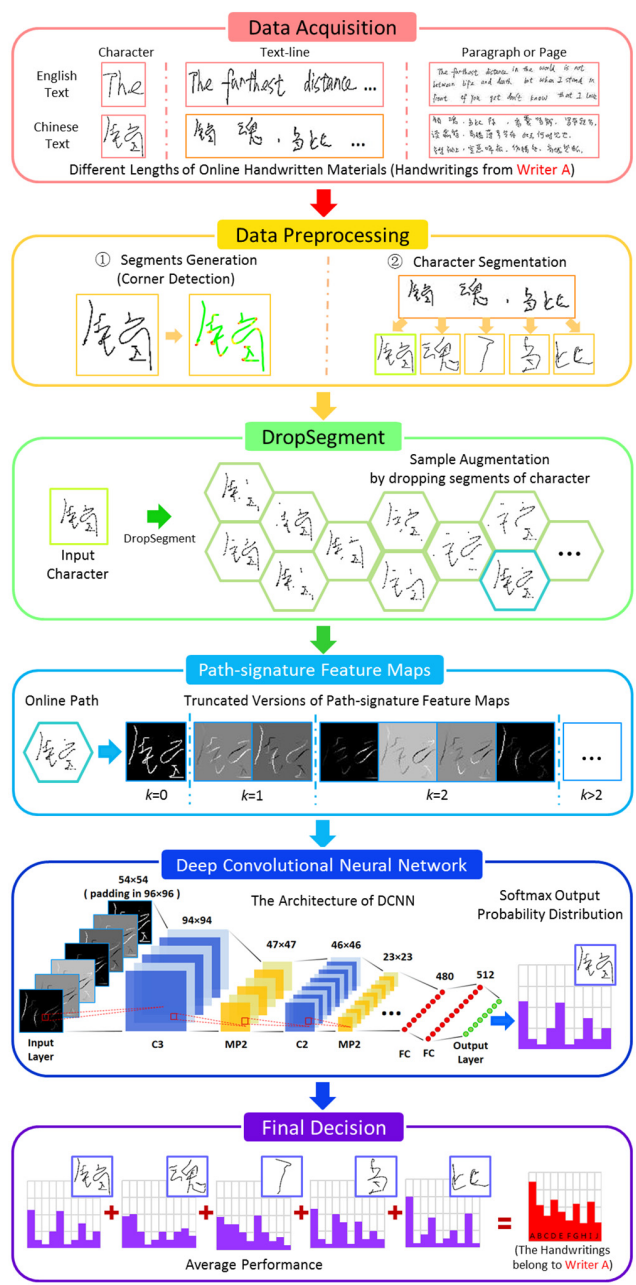

Figure 1. Illustration of DeepWriterID for online handwriting-based writer identification.

## 1. INTRODUCTION

Writer identification is a task of determining a list of candidate writers according to the degree of similarity between their handwriting and a sample of unknown authorship [1]. Currently, it is popular owing to the development and commercialization of touchscreen or pen-enabled electronic devices such as smartphones, and tablet PCs. Its wide range of downstream uses include distinguishing forensic trace evidence, performing mobile bank transactions, and authenticating access to networks. Since most of these applications are closely related to the purpose of assuring personal and property security, handwriting identification merits more attention from academia and industry.

Identifying the handwriting of a writer is one of the highly challenging problems in the fields of artificial intelligence and pattern recognition. Conventionally, handwriting identification systems follow a sequence of data acquisition, data preprocessing, feature extraction, and classification [2]. Research into handwriting identification has been focused on two categories: offline and online. Offline handwritten materials are considered more general but harder to identify, as they contain merely scanned image information. In contrast, systems dealing with online handwriting are popular and expected to achieve better performance since the development of devices that make it possible to acquire handwriting rich in information (e.g., position, velocity, pressure and altitude). Another way to categorize handwriting identification systems is by whether they are text-dependent or text-independent. Text-dependent methods provide high accuracy but are inapplicable in cases where text content is absent, while text-independent methods are robust against content but require a large amount of data to ensure their generalized applicability. In the evaluation stage, different lengths of source materials (e.g., character, text line, page, and document) result in varying levels of difficulty in acquiring sufficient information for identification. In addition, the multiple languages of materials can be evaluated individually or integrally, leading to different requirements for the generalized applicability of the system.

Although numerous researchers have been dedicated to addressing the handwriting identification problem and have achieved tremendous progress [3], this problem is still unsolved in the face of such conditions as insufficiency of data, different lengths of source material, and multiple languages of handwriting material. In this paper, we propose a new method called DropSegment, which randomly removes some number of segments from the characters of an original handwriting sample while retaining the identity information contained in it. DropSegment is a well-performing data augmentation technique as well as an effective way to improve generalized applicability and prevent model over-fitting. In addition, we introduce the path-signature feature to the field of handwriting identification for its ability in extracting discriminant information to uniquely represent a trajectory. Further, we employ a deep convolutional neural network (CNN), which can be regarded as a feature representation method together with a nonlinear classifier, to implement a novel writer identification system. Our system, called DeepWriterID, is presented in Fig. 1.

## 2. DROPSEGMENT METHOD

### 2.1 MOTIVATION

Building a successful handwriting identification system, we often encounter some of the following problems:

**Training data are scarce.** Sufficient handwriting data are necessary not only for text-independent writer identification systems to ensure content-free performance, but also for the deep neural network-based feature representation models to achieve the best performance. However, collecting them is obtrusive and tedious for users in the real world, especially when paragraphs or pages of material are required.

**The generalization capability is insufficient.** The insufficiency of generalized applicability leads to poor system performance, and the use of stroke structure is often blamed. While stroke structure can sometimes be helpful in distinguishing the identity of writers in text-dependent systems, its use limits generalizability when faced with handwriting having diverse structures (e.g., in natural or multilingual handwriting).

**The segmentation problem is vexing.** The lengths of identification materials and the sizes of characters differ in different practical applications. Either over-segmentation or under-segmentation results in different sizes of the characters and adversely affects the identification performance. In addition, when faced with multiple languages, the basic units differ in size (e.g., characters in Chinese and words in English), introducing further difficulties in proper segmentation.

**Ensemble models are costly.** Ensemble methods provide outstanding results, but the storage sizes are too large to employ in practical use, especially for mobile device applications. It is necessary to find a flexible method to achieve similarly successful results while maintaining a constant storage size.

Inspired by Dropout [5], we propose a novel method, called DropSegment, for handwriting identification to alleviate above-mentioned problems.

### 2.2 ANALYSIS

DropSegment is an efficient data augmentation technique. Each original character has at least one stroke, and each stroke contains a certain number of segments, defined by adopting some segmentation methods; for example, we predefine the corner points of a stroke to be its segmentation points. Suppose that an original character has $m$ strokes and that its $i$-th stroke contains $s_i$ ($s_i \in \mathbb{N}^+$) segments. If $d_i$ ($0 \leq d_i \leq s_i$, $d_i \in \mathbb{N}$) segments are dropped from this stroke, then the number of possible combinations of the remaining segments will be

$$C_{s_i}^{d_i} = s_i! / d_i!(s_i - d_i)! . \tag{1}$$

According to the addition principle in combinatorics, the number of all possible combinations derived from the $i$-th stroke is the sum of (1) over all $d_i$, yielding

$$\sum_{d_i=0}^{s_i} C_{s_i}^{d_i} = 2^{s_i} . \tag{2}$$

Then, according to the multiplication principle, the number of new characters generated from the prototype character is the product of (2) over all strokes $i$ ($1 \leq i \leq m$, $i \in \mathbb{N}^+$), expressed as:

$$N(m, S) = \left( \prod_{i=1}^{m} \sum_{d_i=0}^{s_i} C_{s_i}^{d_i} \right) - C_{\hat{s}}^{\hat{s}}, \tag{3}$$
$$= 2^{\hat{s}} - 1$$

where the sequence $S = \{s_1, s_2, \ldots, s_m\}$ contains the number of segments in each stroke of the original character. The sum of the segment counts is $\hat{s} = \sum_{i=1}^{m} s_i$. Since we would never remove all the segments of a character, we exclude the $C_{\hat{s}}^{\hat{s}}$ in (3). Two examples are given using DropSegment: a 3-stroke character with $S = \{2, 3, 4\}$ can generate 511 new characters (calculated by $(2^2)^1 \times (2^3)^1 \times (2^4)^1 - 1$), and an 8-stroke character with all $s_i = 3$ ($i = 1, 2, \ldots, 8$) can generate more than $1.67 \times 10^7$ new characters, calculated by $(2^3)^8 - 1$ based on (3). Therefore, with some number of segments being randomly dropped from the characters, the remaining segments recombine to form a massive number of new characters that appear diverse, thereby achieving data augmentation. Examples of new characters together with their prototype character are shown in Fig. 1.

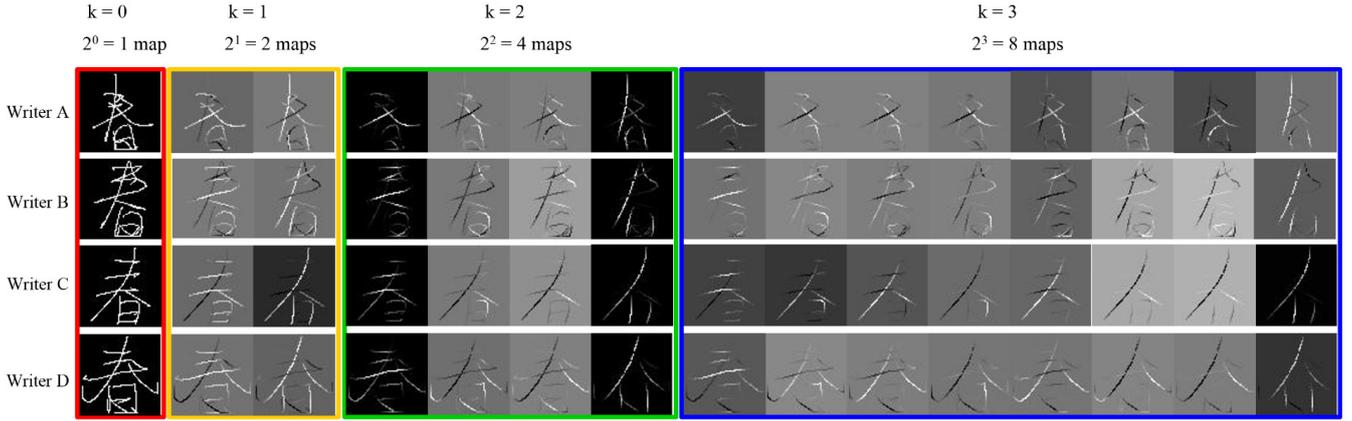

Figure 2. Visualization of the path-signature feature. The parameter $k$ denotes the $k$-th–level iterated integral of the path-signature.

In practical handwriting identification, the stroke structure of various source materials (e.g., in multiple languages, ranging from neat handwriting to scrawled strokes) is no longer an invariant and may even be harmful to the identification. By separating strokes, DropSegment can prevent the structural information from being considered, and maintain the writer discriminant information and thus improve its generalized applicability. Additionally, DropSegment is robust to over-segmentation and under-segmentation problems. When DropSegment is adopted, all the characters will have their segments omitted probabilistically, so the segmentation method can be relatively crude. Moreover, in the testing stage, our approach elastically generates a certain number of new test samples to provide multiple predictions. Averaging these predictions is expected to improve performance with no extra consumption of storage.

Compared with DropStroke [7] which removes strokes from a character, DropSegment is more robust for three reasons. First, in fast text-line handwriting or cursive handwritten personal signatures, strokes that are formally separated may be concatenated, so wiping off the whole stroke can lead to too much information loss. Second, the basic units in some languages have scarce strokes, such as English letters, accounting for the limited generalizability of the DropStroke method [7] when faced with multilingual handwriting. Third, since segments are structures that are more detailed than strokes, DropSegment can generate more possible samples than DropStroke. Notice that in (3), whenever $d_i = s_i$, the effect of DropSegment is equivalent to that of DropStroke [7], accounting for the generalized applicability of DropSegment.

## 3. GENERATION OF PATH-SIGNATURE FEATURE MAPS

The path-signature, pioneered by Chen [8] in the form of the collection of iterated integrals and developed in recent years by Lyons to play a fundamental role in rough theory [9,10,11], is able to extract a sufficient quantity of information concealed in a path of finite length (e.g., speech traces or online handwriting) to solve any linear differential equation. For handwriting analysis, the path-signature feature was first introduced by Graham [4] to address handwritten characters recognition. In DeepWriterID, we sought a way to extract further valuable information for the handwriting identification problem using path-signature feature maps.

Given a pen segment $P$ of finite length writing in the plane $\mathbb{R}^2$, we can denote a continuous mapping $P:[0,T] \to \mathbb{R}^2$. Letting $k \in \mathbb{N}$ and $0 < \tau_1 < ... < \tau_k < T$, the $k$-th–level iterated integral of path $P$ can be represented as

$$P_{0,T}^k = \int_0^T \int_0^{\tau_k} ... \int_0^{\tau_2} dP_{\tau_1} \otimes \cdots \otimes dP_{\tau_k} . \quad (4)$$

The dimension of $P_{0,T}^k$ is $2^k$. For algebraic reasons, when $k = 0$, the signature is constant at 1, denoting the original input, while $k = 1$ and $k = 2$ represent the path displacement and path curvature, respectively. When $P$ is a straight line, the iterated integrals $P_{0,T}^k$ can be calculated iteratively by:

$$P_{0,T}^k = \begin{cases} (P_{0,T}^{k-1} \otimes \Delta_{0,T})/k & k \geq 1 \\ 1 & k = 0 \end{cases}, \quad (5)$$

where $\Delta_{0,T}$ denotes the path displacement. By increasing the number of iterations of the integrals, higher levels of path information can be revealed, but the dimension of the feature (calculated as $2^k$ for the $k$-th level of the signature) increases exponentially as well. Therefore, to enrich the path representation while keeping the computational time for features to a minimum, the collection of the signature, which means the combining of different levels of integral iterations, is our solution. It is expressed by:

$$F_{0,T}^n = (P_{0,T}^0, P_{0,T}^1, P_{0,T}^2, ..., P_{0,T}^n), \quad (6)$$

where $n$ ($n \in \mathbb{N}$) is the level at which the signature is truncated. At this stage, each sample point along the segment can generate a set of signature values with the truncated level $n$. The dimension of (6) (i.e., the number of feature maps) can be calculated as $2^{n+1} - 1$.

To demonstrate feature maps intuitively, we assign the pixels of the pen trajectory with the corresponding signature image histogram equalization for each feature map, the visualization is shown in Fig. 2. The rows denote different writers, and the columns present the maps of different levels of path-signature feature. We observe that the maps of the first three iterated integrals seem similar in magnitude among the writers, while the differences between the maps beyond the second level ($k > 2$) are appreciable (e.g., the third level). This supports our hypothesis that higher levels of the path-signature may contain information that will contribute positively to handwriting identification.

## 4. ARCHITECTURE AND CONFIGURATION OF DEEP CNN

To achieve integrated optimization in the overall pipeline, we employ a deep convolutional neural network (DCNN) model that exploits the spatial sparsity of handwriting, as described in [4]. The idea of spatial sparsity derives from the online trajectory's containing scarce foreground pixels compared with the numerous background pixels, which take values of zero, so the background computation can be avoided to save time. As shown in Fig. 1, the architecture of our DCNN includes 5 convolutional layers, each of which is accompanied by a max-pooling (MP) layer. We fix the size of the convolutional filter at 3×3 (denoted by C3) for the first layer and 2×2 (C2) for the others, with a stride of 1 pixel. The window size of max-pooling is 2×2 (MP2) with a stride of 2 pixels. The small filter size and pooling size enable the network to retain the valuable information inside. At the top of our network, two fully connected (FC) layers with 480 and 512 units in size, respectively, are included in the design in order to better characterize the complicated biometric information and to provide additional nonlinearity to the network. The number of convolutional filter kernels is 80 for the first layer and is incremented by 80 after each max-pooling. For activation functions, rectified linear units (ReLUs) are used for neurons in the convolutional layers and FC layers, and softmax is used for the output layer.

Our DCNN renders the input data into a 54×54 bitmap and puts it in the center of a 96×96 grid, where the extra pixels beyond the bitmap are the budget of padding pixels for all borders in the convolutional layers. The architecture of our networks can be uniformly represented by:

M×96×96Input-80C3-MP2-160C2-MP2-240C2-MP2-320C2-MP2-400C2-MP2-480FC-512FC-Output,

where M denotes the number of input channels, which is equal to the number of signature feature maps.

## 5. EXPERIMENTAL RESULTS

### 5.1 DATABASE

We used the National Laboratory of Pattern Recognition (NLPR) handwriting database [12]. It contains four pages of Chinese text from each writer, and four pages of English text from each writer. To evaluate the performance of DeepWriterID, we performed experiments on two subsets: Dataset I (DB I), which was contributed by 187 writers, includes two free-content Chinese pages for training and one fixed-content Chinese page for testing; and Dataset II (DB II), which was contributed by 134 writers, includes two free-content English pages for training and one fixed-content English page for testing.

The samples in the NLPR database were collected by a Wacom Intuos2 tablet and include rich sequential information including pen-position, pen-down and pen-up states, azimuth, altitude, and pressure. In general, the only consistently available information on most touchscreen mobile devices is the pen-position of handwriting in the pen-down state; thus, we deliberately ignored the other handwriting information and conducted our experiments using pen-position in the pen-down state.

### 5.2 DATA PREPROCESSING

The samples in the NLPR database [12] are presented on pages, and the handwriting can be regular or cursive, so the segmentation is employed as a data preprocessing step in DeepWriterID in order to unify the various input data. The segmentation is twofold: segments generation and pseudo character segmentation.

**Segments generation.** For each stroke on a page, we employed a fast and efficient corner detection algorithm [6] for segments generation. To detect corners, this algorithm employs the concept that the directions of the forward and backward vectors of a non-corner point will cancel each other. A bending value is defined as

$$\beta = \max\left( |x_{i+k} + x_{i-k} - 2x_i|, |y_{i+k} + y_{i-k} - 2y_i| \right)/2k, \quad (7)$$

where $(x_i, y_i)$ are the trajectory points after interpolation, $(x_{i+k}, y_{i+k})$ and $(x_{i-k}, y_{i-k})$ are the corresponding $k$-th forward and backward points, respectively. The value of $k$ is set as 2 in this paper. Bending values assess the degree of curvature and are calculated for each point on the trajectory, and the local maximum bending values are defined to be the corners. Then, each stroke is divided into different segments according to these corners. The heat map of bending values for a sample pen trajectory is illustrated in Fig. 3.

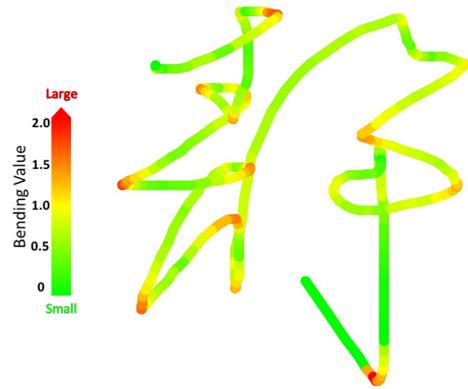

Figure 3. A heat map of bending values. The color red on the trajectory indicates a large bending value, while green denotes a relatively small value. The points having local maximum bending values are considered as potential segmentation points.

**Pseudo character segmentation.** After all the strokes have been separated into segments, the character segmentation follows, using the assumption that the width of a character can be roughly estimated by its height. In fact, we do not need an accurate segmentation of character for the writer identification task, therefore we apply a pseudo character segmentation process without loss the discriminant information of different writers. To unify the size of CNN inputs, the height-to-width ratio should be the same. We thus determined this ratio to be 1 for simplicity. From experiments, we found that other ratio value between [1, 1.3] is also proper to produce similar result. During the pseudo character segmentation, the average height is measured first. Then, each of the segments will be sequentially assigned to form a pseudo character. When the width of a character exceeds the average height after adding a segment, this segment is regarded as the beginning of a new character. It is worth to note that the pseudo characters formed by this way are not always meaningful, but uniformly similar to square in shape. Since the input samples required for the writer identification task are independent of their meanings, it is not essential to employ accurate segmentation according to meaningful characters. The samples generated by this way may contains non-character or may consists of 2 ~ 3 letters (for English text) instead of a single letter or a word, but we found that such samples do not prevent us to use them as useful and effective CNN training data for writer identification.

### 5.3 DROPSEGMENT IMPLEMENTATION

After the data preprocessing, all the data on pages have been separated into characters with segments. In our implementation, for each iteration, we first randomly select a mini-batch of these characters. If a character contains $m$ ($m \in \mathbb{N}^+$) segments, a random number $r$ ($r \in [0, \lfloor m/2 \rfloor], r \in \mathbb{N}$) of the $m$ segments will be removed; the removed segments are also randomly chosen. Our DropSegment using this setting not only generates a considerable quantity of samples but also prevents too much information loss. For the two examples mentioned in Section 2.2, the numbers of generated characters reduce to 256 and $9.74 \times 10^6$, respectively.

Notice that the remaining segments appear fragmented and may destroy the writing styles, so combining the segments is necessary. For each character, after removal of segments, the remaining segments that are concatenated with each other in the prototype character will be recombined to form longer segments. After that, each isolated segment will individually become a new stroke. Hence, DropSegment can maintain most of the identity information while achieving data augmentation.

### 5.4 INVESTIGATION OF PATH-SIGNATURE FEATURE

In our experiment, we employed the single DCNN as described in Section 4. During the training stage, before the extraction of path-signature feature maps, we adopted a random mix of affine transformations (i.e., translation, rotation, scaling) as the elastic distortion to achieve further data augmentation and generalized applicability. The mini-batch size was set to 100, and the Dropout [5] rates for the last four layers were empirically set to 0.3, 0.4, 0.5, and 0.5, respectively. We spent a week training the DCNN on a PC with a GTX 980 graphics processing unit. The DCNN was trained on character-level samples, and noting that the softmax output of the network can be treated as a probability distribution of all classes, we thus averaged the outputs of all the characters from each page to give the final prediction at the page level, as illustrated in Fig. 1.

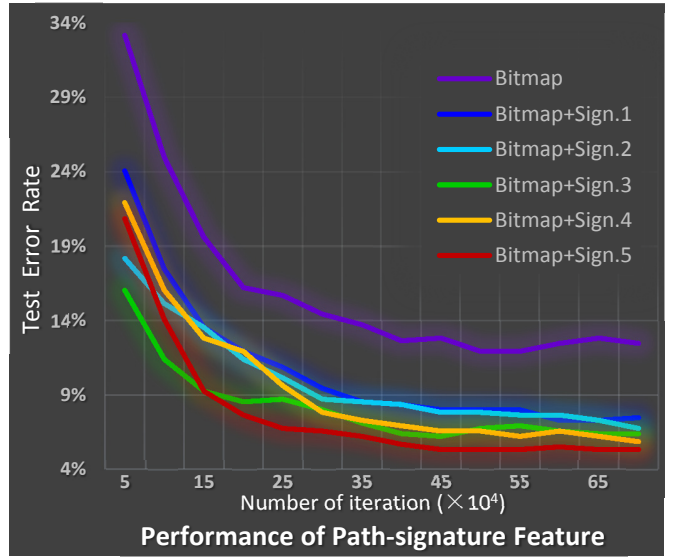

Figure 4. Performance of different truncated levels of path-signature feature. The page-level results shown are the Top-1 test error rates on DB I (the Chinese dataset).

The baseline method (denoted by "Bitmap") in our experiments is to render the online handwritten characters as offline bitmaps to train the DCNN. We found that a DCNN with additional path-signature features can incorporate prior knowledge in the representation and significantly improve the performance. Thus, we evaluated the effect of the path-signature feature on handwriting identification using the Chinese in DB I. The curves of page-level test error rate are shown in Fig. 4. The path-signature feature at truncated level n is denoted by Sign.n. We discovered that the path-signature feature together with Bitmap can greatly enhance the network representation and beat the baseline by a large margin. The higher truncated levels of signature provides more subtle handwriting information, accounting for their better results, but the improvements become more negligible with the exponentially increasing dimensions of feature.

### 5.5 INVESTIGATION OF DROPSEGMENT METHOD

Next, we conducted experiments on Chinese text in DB I to evaluate the proposed DropSegment method. DropSegment is flexible and produces varying results as it randomly generates a certain number of new test samples from the prototypes. Therefore, it makes sense to combine these results to achieve better performance without re-training new models. We averaged 20 of these predictions at the test stage in our experiments. The page-level performance is shown in Table 1. Notice that with DropSegment, all the identification rates are markedly improved over those without DropSegment.

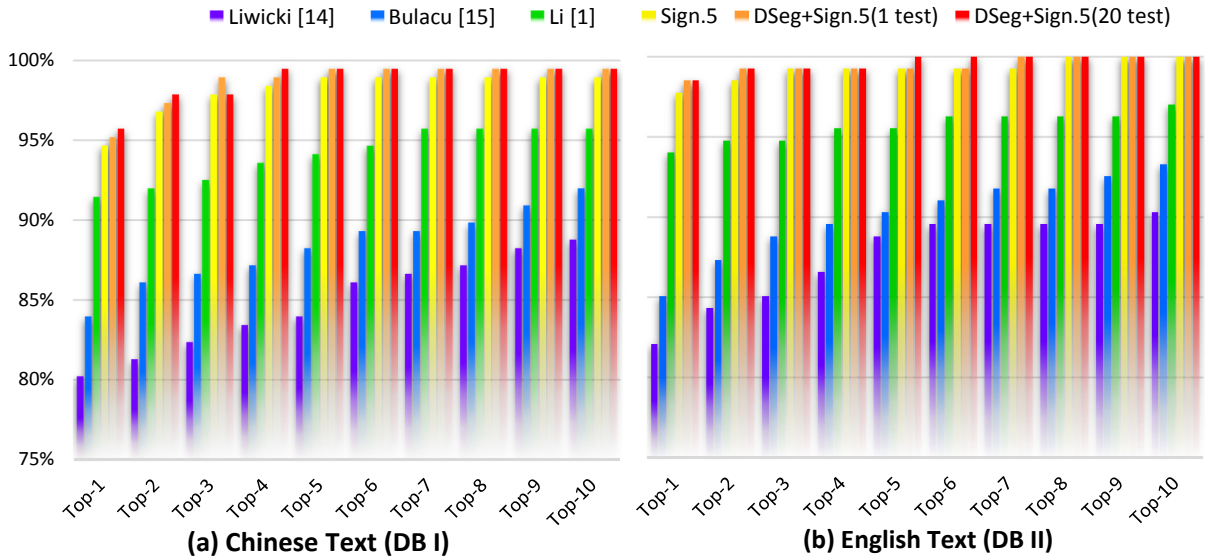

Figure 5. The Top-10 average writer identification rates of different methods on DB I (a) and DB II (b).

Table 1. Average writer identification rates (%) of different methods with and without DropSegment on Chinese Text (DB I).

| Methods | Without (1 test) | With DropSegment (1 test) | With DropSegment (20 test) |
|---|---|---|---|
| **Baseline**:Bitmap | 87.70 | 89.30 | **89.84** |
| Bitmap+Sign.1 | 91.98 | 93.05 | **93.05** |
| Bitmap+Sign.2 | 92.51 | 93.05 | **93.58** |
| Bitmap+Sign.3 | 93.05 | 94.12 | **94.65** |
| Bitmap+Sign.4 | 94.12 | **94.65** | 94.12 |
| Bitmap+Sign.5 | 94.65 | 95.19 | **95.72** |

A comparison of three previous methods and ours is shown in Fig. 5. The first method, introduced by Liwicki et al. [13], combines features with point-level and stroke-level information. The second method, proposed by Bulacu and Schomaker [14], uses both texture-level and allograph-level features and evaluates offline samples. The third method, which previously produced the best results with the NLPR database [12], employs hierarchy models with features extracted by the shape primitives probability distribution function. The best results previously reported were 91.50% on DB I (Chinese text) and 93.60% on DB II (English text), respectively [1]. These results were mostly achieved by combining sufficient features extracted from rich online information including pressure, altitude, azimuth, velocity, and pen-position in both pen-down and pen-up states. However, our approach attained the highest writer identification rates of 95.72% for Chinese text and 98.51% for English text, indicating relative error reduction rates of 49.6% and 76.5%, respectively, and this was achieved even though we only use pen-position information.

## 6. CONCLUSIONS

By introducing DropSegment, the path-signature feature, and deep CNN, we have designed a new approach, called DeepWriterID, to address the online text-independent writer identification problem. The path-signature feature we introduced to handwriting identification has proved to be very useful and effective. The proposed DropSegment technology makes it possible to train excellent CNNs even when data are inadequate, as in the case of writer identification. Furthermore, it has the benefit of achieving promising ensemble performance without training or needing to store additional network models. Finally, DeepWriterID achieves new state-of-the-art writer identification accuracies on two subsets of the NLPR database.

It is worth to note that the proposed method does not consider rejection of unknown writers, which is an important issue for further study. One possible way to reject unknown writers can be done by analyzing the confidence measurement learnt by CNN. Moreover, DeepWriterID should also allow registration of new writers that are not included in the training database, as this would provide a flexible way to absorb new characteristics and offer a means of online or incremental learning, which is also worth studying in the future.

**ACKNOWLEDGEMENT**

This research is supported in part by NSFC (Grant No.: 61472144), GDSTP (Grant No.: 2014A010103012, 2015B010101004) and GDUPS (2011).